\PassOptionsToPackage{dvipsnames}{xcolor}
\documentclass[letterpaper, 10 pt, conference]{ieeeconf}  

\IEEEoverridecommandlockouts                              

\usepackage{cite}
\usepackage{amsmath,amssymb,amsfonts}
\usepackage{algorithmic}
\usepackage{textcomp}
\usepackage{graphicx}
\usepackage{xcolor}
\usepackage{makecell}
\usepackage{multirow}
\usepackage{booktabs}
\usepackage{float}
\usepackage{svg}
\usepackage{stfloats}

\let\labelindent\relax
\usepackage{enumitem}

\definecolor{mycolor}{RGB}{100, 150, 200} 

\title{\LARGE \bf SparseMeXt: Unlocking the Potential of Sparse Representations for HD Map Construction
}

\begin{document}

\author{
    Anqing Jiang$^{1*}$, 
    Jinhao Chai$^{1,2}$, 
    Yu Gao$^{1}$, 
    Yiru Wang$^{1}$, 
    Yuwen Heng$^{1}$, 
    Zhigang Sun$^{1}$,
    Hao Sun$^{1}$, \\
    Zezhong Zhao$^{3}$, 
    Li Sun$^{3}$,
    Jian Zhou$^{3}$,
    Lijuan Zhu$^{1}$,
    Hao Zhao$^{4}$
    Shugong Xu$^{5}$
    \thanks{$^{1}$
            Anqing Jiang,
            Yuwen Heng,
            Yu Gao,
            Yiru Wang,
            Zhigang Sun,
            Hao Sun are with Bosch Corporate Research, Bosch (China) Investment Ltd., Shanghai, China}
    \thanks{$^{2}$
        Jinhao Chai,
        is with School of Communication and Information Engineering, Shanghai University, Shanghai, China
    }
    \thanks{$^{3}$
        Zezhong Zhao, 
        Li Sun,
        are with Bosch Mobility Solutions, Robert Bosch GmbH, Suzhou, China
    }
    \thanks{$^{4}$ 
        Hao Zhao,
        is with AIR, Tsinghua University, Beijing, China
    }
    \thanks{$^{5}$ 
        Shugong Xu,
        is with Xi'an Jiaotong-Liverpool University, Suzhou, China
    }
}

\maketitle
\pagestyle{empty}

\begin{abstract}

Recent advancements in high-definition (HD) map construction have demonstrated the effectiveness of dense representations, which heavily rely on computationally intensive bird’s-eye view (BEV) features. While sparse representations offer a more efficient alternative by avoiding dense BEV processing, existing methods often lag behind due to the lack of tailored designs. These limitations have hindered the competitiveness of sparse representations in online HD map construction. In this work, we systematically revisit and enhance sparse representation techniques, identifying key architectural and algorithmic improvements that bridge the gap with—and ultimately surpass—dense approaches. We introduce a dedicated network architecture optimized for sparse map feature extraction, a sparse-dense segmentation auxiliary task to better leverage geometric and semantic cues, and a denoising module guided by physical priors to refine predictions. Through these enhancements, our method achieves state-of-the-art performance on the nuScenes dataset, significantly advancing HD map construction and centerline detection. Specifically, SparseMeXt-Tiny reaches a mean average precision (mAP) of 55.5\% at 32 frames per second (fps), while SparseMeXt-Base attains 65.2\% mAP. Scaling the backbone and decoder further, SparseMeXt-Large achieves an mAP of 68.9\% at over 20 fps, establishing a new benchmark for sparse representations in HD map construction. These results underscore the untapped potential of sparse methods, challenging the conventional reliance on dense representations and redefining efficiency-performance trade-offs in the field.

\begin{figure}[h!] 
    \centering
    \includegraphics[width=0.5
    \textwidth]{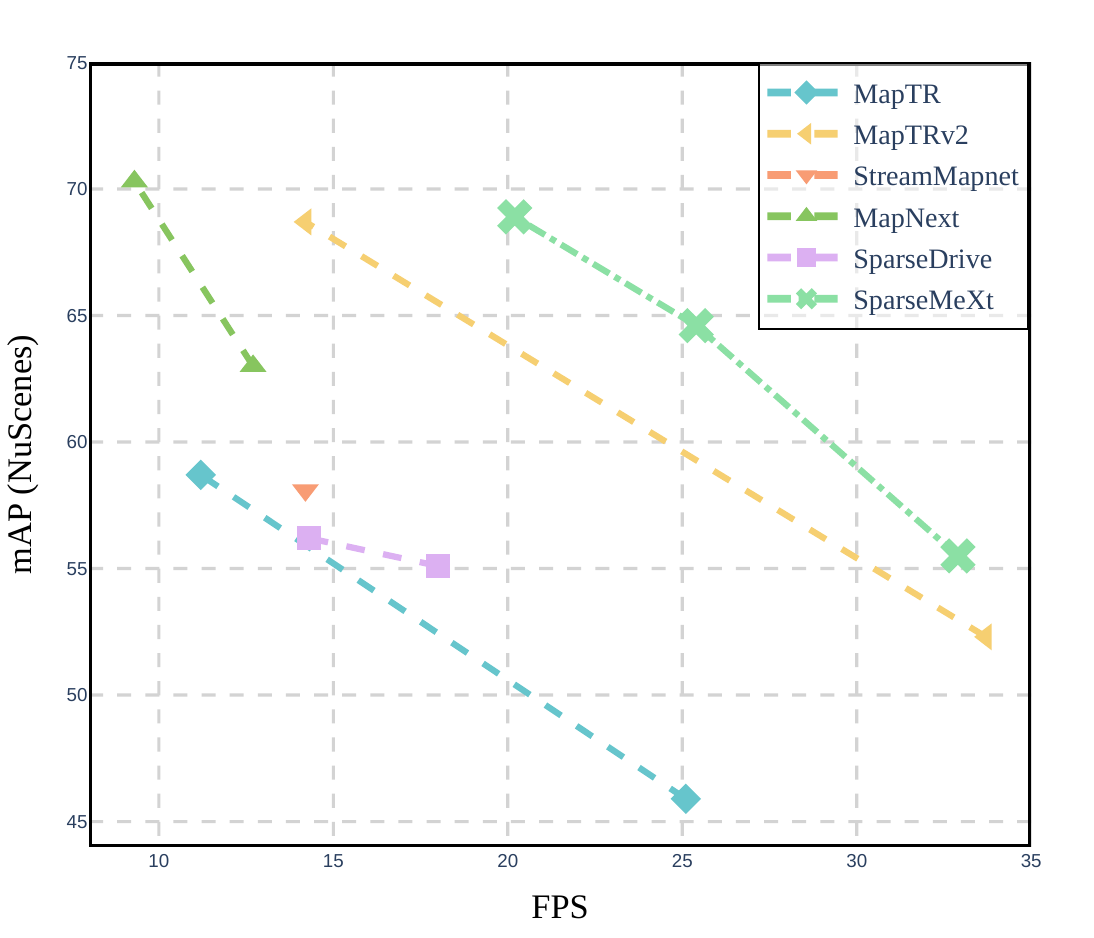} 
    \caption{This figure illustrates the trade-off between FPS (frames per second) and mAP (mean average precision) on the nuScenes dataset. SparseMeXt achieves a superior balance between efficiency and accuracy, significantly outperforming other sparse methods like SparseDrive, while even surpassing dense approaches such as MapTRv2. These results underscore the untapped potential of sparse representations in HD map construction. FPS values are measured on a NVIDIA RTX 3090.}
    \label{fig:my_label} 
\end{figure}
\end{abstract}

\section{INTRODUCTION}

In autonomous driving, the environment is composed of both dynamic and stationary entities. Dynamic entities, such as vehicles and pedestrians, are constantly moving and interacting, while stationary entities, including road markings, traffic signs, and lane boundaries, serve as critical infrastructure for traffic regulation. High-definition (HD) maps play a vital role in autonomous navigation by providing precise geometric and semantic information about these stationary elements. Unlike offline maps, online HD maps dynamically update in real-time, offering up-to-date environmental data that is essential for vehicle localization, planning, and decision-making, ultimately ensuring safer and more efficient autonomous operation.

Traditional HD map construction has long relied on manual annotation and SLAM-based methods \cite{zhang2014loam, shan2018legoloam, bai2022fasterlio, xue2022legoloamsc, zheng2022fastlivo}, which are both costly and difficult to maintain in rapidly changing environments. To improve scalability, recent real-time HD map construction approaches \cite{reinauer2021persformer, zhou2022crossviewlane, can2021structured} leverage line-shape priors to detect lanes directly from front-view images. With the rise of Bird’s-Eye-View (BEV) representations \cite{huang2021bevdet, li2022bevformer, li2023bevdepth}, HD map construction has shifted towards learning-based methods. Among these approaches, HDMapNet \cite{li2022hdmapnetonlinehdmap} generates vectorized maps via pixel-wise segmentation, while VectorMapNet \cite{liu2023vectormapnetendtoendvectorizedhd} represents map elements as point sequences. The MapTR series \cite{liao2023maptrstructuredmodelinglearning, liao2024maptrv2endtoendframeworkonline} adopts a DETR-like paradigm \cite{carion2020endtoendobjectdetectiontransformers} for efficient vectorized map construction. However, these methods heavily rely on dense perspective transformation modules, leading to computational costs that scale with perception range, making deployment on low-power or resource-constrained devices challenging.

To address this, the Sparse4D series introduces a deformable 4D aggregation module as an alternative to dense BEV features. SparseDrive \cite{sun2024sparsedriveendtoendautonomousdriving}, built on Sparse4D, pioneers fully sparse scene representations for HD map construction, significantly improving efficiency. However, its performance still lags behind dense approaches. To bridge this performance gap, we systematically analyze sparse network designs, identifying key architectural limitations and proposing three novel enhancements that significantly improve HD map construction performance using sparse representations.

First, we notice that object detection tasks exhibit dynamic image occupancy, where distant vehicles occupy small regions while nearby ones cover larger areas. In contrast, HD map construction involves consistently large spatial coverage across the image. This fundamental difference suggests that while cascade feature aggregation is effective for sparse 3D object detection, it is inefficient for map-related tasks. Existing architectures, originally designed for 3D object detection, lack tailored optimizations for sparse map feature extraction. To address this, we introduce a network architecture specifically optimized for map-related features.


Second, dense paradigms commonly employ semantic segmentation auxiliary supervision \cite{li2022bevformer, chen2022persformer3dlanedetection, liu2022petrv2unifiedframework3d} to refine BEV feature generation for 3D object detection and HD map construction. However, in sparse detection tasks, this approach fails due to the lack of explicit BEV feature grids, leading to the loss of global contextual information. To mitigate this, we propose a sparse-dense instance-to-segmentation auxiliary task, which enhances supervision by better leveraging semantic and geometric information. 


Third, as a DETR-like detector, SparseDrive suffers from query inconsistency, where the same input image results in query mismatches across different training epochs. This inconsistency introduces optimization ambiguity, hindering convergence and reducing overall performance. Moreover, its denoising training strategy, originally designed for box-shaped object detection, struggles to handle the diverse geometric forms of HD map elements, such as point sequences. Inspired by DN-DETR\cite{li2022dn}, we design an optimized denoising module guided by physical priors, incorporating line shifting, angular rotation, scale transformation, and curvature adjustments. By ensuring that noise perturbations adhere to physical constraints, our approach significantly improves the effectiveness of denoising strategies for map tasks. 



To the best of our knowledge, our SparseMeXt is the first sparse paradigm to surpass dense methods in online HD map construction, setting a new benchmark in the field. The key contributions of our work are summarized as follows:

\begin{itemize}
\item We introduce a network architecture tailored for sparse map feature extraction, addressing the inefficiencies of existing object detection-centric designs in HD map construction. Our approach optimizes feature aggregation and representation learning to better capture the large-area coverage required for map-related tasks.
\item We introduce a sparse-to-dense auxiliary segmentation supervision method to compensate for the absence of BEV features in sparse paradigms. By integrating instance-level supervision with scene-level segmentation tasks, this approach effectively captures both semantic and geometric information.
\item We propose a physical prior-based query denoising strategy for online HD map construction. This includes a detailed framework for generating various physical prior noise types, along with the incorporation of temporal priors, to enhance prediction stability and robustness.
\item Designed for real-time onboard deployment, SparseMeXt achieves a 10.1\% mAP improvement over SparseDrive, demonstrating substantial performance gains. Additionally, compared to strong baseline MapNext in dense paradigms, SparseMeXt surpasses 2.2\% mAP, further solidifying the competitiveness of sparse representations in HD map construction.
\end{itemize}

\section{Related Work}
\subsection{HD map construction}

Recent advancements in 2D-to-BEV methods \cite{li2022bevformer} have reframed HD map construction as a segmentation task, leveraging surround-view image data from vehicle-mounted cameras to produce rasterized maps via BEV semantic segmentation. For vectorized HD map, early approaches such as HDMapNet \cite{li2022hdmapnetonlinehdmap} depend on heuristic post-processing to aggregate pixel-wise segmentation results, a process that is both time-consuming and computationally inefficient. In contrast, VectorMapNet \cite{liu2023vectormapnetendtoendvectorizedhd} introduces the first end-to-end framework, employing a two-stage coarse-to-fine strategy with an autoregressive decoder. However, this approach encounters challenges, including slow inference speeds and permutation ambiguities. To overcome these limitations, MapTR \cite{liao2023maptrstructuredmodelinglearning} proposes a unified one-stage framework that resolves ambiguities and enhances efficiency. Its successor, MapTRv2 \cite{liao2024maptrv2endtoendframeworkonline}, further improves performance by incorporating an auxiliary one-to-many matching branch and utilizing depth supervision to better capture 3D geometric information. More recently, StreamMapNet \cite{yuan2024streammapnet} integrates temporal information through a streaming strategy, markedly enhancing the temporal consistency and quality of vectorized HD map. Collectively, these advancements elevate the accuracy, efficiency, and robustness of online HD map construction. However, these methods require maintaining a dense BEV feature space, which imposes high computational complexity, limiting their adaptability to larger BEV ranges or deployment on low-computational platforms.

\subsection{3D Sparse Perception}
Early object detection methods\cite{xu2021confine, liu2016ssd, ren2016faster, tan2020efficientdet, tian2022fullyconvolutional, sunzhigang10592819} used dense predictions as output, and then utilized non-maxima suppression (NMS) to process those dense predictions. DETR \cite{carion2020endtoendobjectdetectiontransformers} introduces a new detection paradigm that utilizes set-based loss and transformer to directly predict sparse detection results. DETR3D\cite{wang2021detr3d3dobjectdetection} is a representative work of sparse methods, which performs feature sampling and fusion based on sparse reference points. Sparse4D series\cite{lin2022sparse4dv1, lin2023sparse4dv2, lin2023sparse4dv3} can efficiently and effectively achieve 3D detection without relying on dense view transformation nor global attention, and is more friendly to edge devices deployment, and extend the detector into a tracker using a straight forward approach that assigns instance ID during inference, further highlighting the advantages of query-based algorithms. SparseDrive\cite{sun2024sparsedriveendtoendautonomousdriving} and SparseAD\cite{zhang2024sparsead} include detection, tracking and online mapping are unified with general temporal decoders\cite{wang2023objectcentrictemporal}, in which multi-sensor features and historical memories are regarded as tokens, and object queries and map queries represent obstacles and road elements in the driving scenario respectively. However, the task of online HD map construction using sparse architectures remains underexplored, with limited systematic study contributing to lower performance compared to dense BEV approaches.


\begin{figure*}[t!] 
    \centering
    \includegraphics[width=1.0\textwidth]{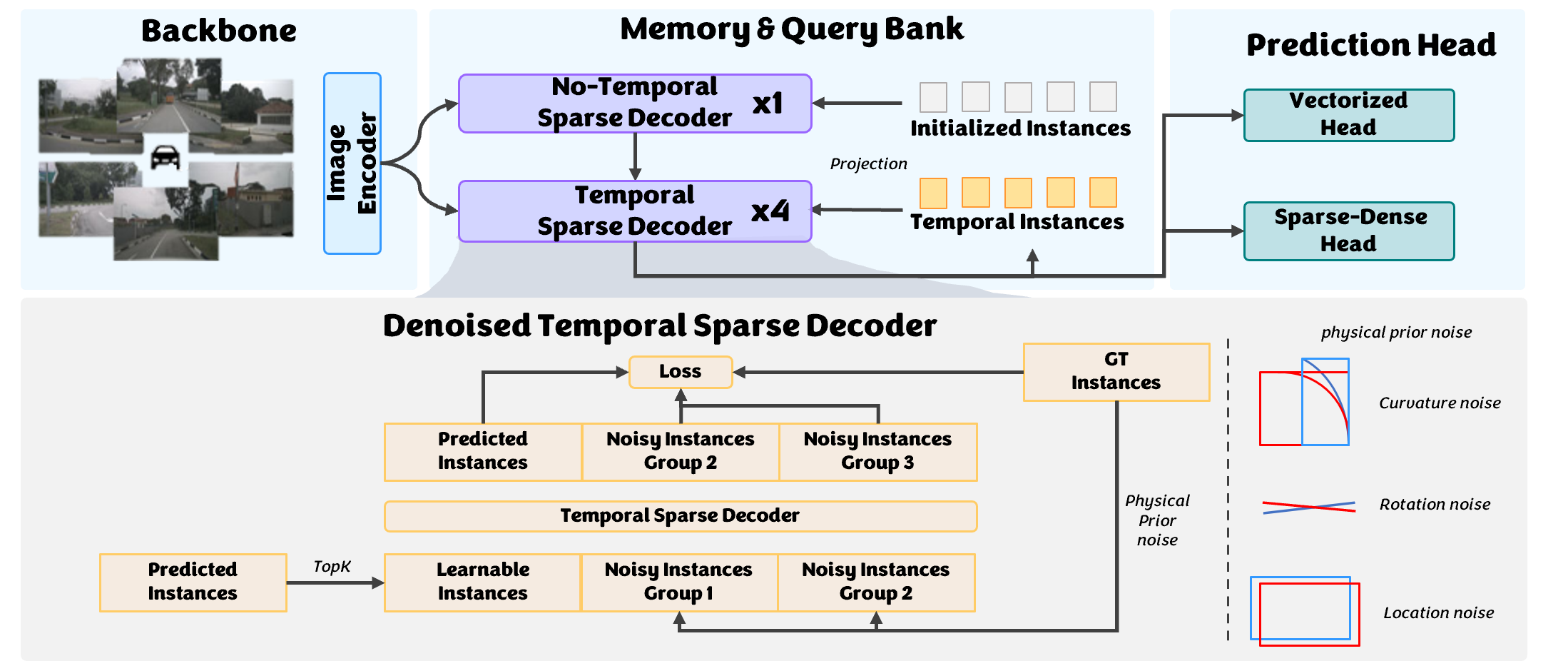} 
    \caption{Overall architecture of SparseMeXt. }
    \label{fig:Overview} 
\end{figure*}

\section{Method}

In this section, we first provide a short summary of our SparseMeXt. Subsequently, we decompose the model architecture and analyze its components in detail.


\subsection{Framework Overview}
As illustrated in Fig.\ref{fig:Overview}, SparseMeXt follows a sparse encoder-decoder framework designed for map vectorization. The image encoder, which consists of a backbone and a neck structure, extracts image features. A query memory bank is introduced to facilitate temporal modeling, while the denoised temporal sparse map decoder incorporates a map denoising task by injecting physical prior noise. Finally, we describe how to better leverage semantic and geometric information is integrated through a auxiliary segmentation task.

\subsection{Architecture Optimization}
\subsubsection{Modern Image Backbone}
SparseDrive and MapTR initialize their ResNet backbone with ImageNet pre-trained weights, a traditional approach for transfer learning. However, our pilot study reveals that ImageNet may not be an ideal source dataset for pre-training in our context. This is primarily because ImageNet's object-centric images largely deviate from driving scenarios. Additionally, ImageNet pre-trained networks are optimized for classification tasks, requiring significant adaptation for downstream tasks, especially those sensitive to localization. To address this domain gap, we initialize ResNet-50 with weights from the DD3D depth dataset. Unfortunately, we do not achieve an improvement.
Interestingly, such pre-training settings have been reported to significantly enhance camera-based BEV 3D object detection. Based on these observations, we emphasize the importance of selecting a highly relevant task and dataset for backbone pre-training to minimize the domain gap, as demonstrated in Table~\ref{table:backbone}. We adopt a ResNet-50 backbone pre-trained on nuImages with Cascade R-CNN and achieve a further 1.5\% improvement in mAP, underscoring the benefits of task-relevant pre-training. 


\begin{table}[h!]
\centering
\caption{Performance comparison of backbones with different pre-training datasets on Nuscenes \texttt{val} set.}
\label{table:backbone}
 \begin{tabular}{c|c|cccc} 
 \toprule
\multirow{2}*{Backbone} & \multirow{2}*{Pretrain} & \multicolumn{4}{c}{$AP$} \\
& & $ped.$ & $div.$ & $bou.$ & $avg.$ \\
 \midrule
 R50 & ImageNet cls & 59.1 & 64.3 & 63.4 & 62.3 \\ 
 R50 & DD3D depth & 57.9 & 62.7 & 62.7 & 61.1  (\textcolor{Red}{-1.2})\\
 R50 & nuImages det & \textbf{59.29} & \textbf{66.7} & \textbf{65.4} & \textbf{63.8} (\textcolor{OliveGreen}{+1.5})\\
 
 \bottomrule
 \end{tabular}

\end{table}

\subsubsection{Rethink image encoder neck in HD map construction task}


Previous research \cite{jin2022lookobjects} suggests that, beyond multi-scale feature fusion and divide-and-conquer strategies, the back-propagation paths modified by feature pyramid network (FPN) , which encodes multi-scale features from the backbone and provides feature representations for the decoder by Multiple-in-Multiple-out (MiMo) structure, significantly impact the performance of detection frameworks. This modification restricts each level of the backbone network to detecting objects within a specific scale range, leading to inconsistent changes in the average precision of pedestrian crossing, lane divider, road boundaries ($AP_{\text{ped.}}$, $AP_{\text{div.}}$, and $AP_{\text{bou.}}$). Additionally, YOLOF \cite{chen2021lookonelevelfeature} demonstrates that the primary advantage of FPN lies in its divide-and-conquer approach addressing optimization challenges in dense object detection, rather than its capacity for multi-scale feature fusion. 

Compare to 3D object detection tasks, map elements typically occupy a larger portion of the image. In other words, map element detection more closely resembles large object detection tasks. Inspired by these insights, we adapt a single-in-multiple-out (SiMo) in YOLOF, which only has one input feature Resnet50-C5 and does not perform feature fusion, image encoder neck structure in our SparseMeXt framework. As Table~\ref{table:FPN} shows, SiMo achieves a further 4.25\% improvement in mAP, highlighting the benefits of SiMo image encoder neck.

\begin{table}[h!]
\centering
\caption{The performance comparison of different image encoder neck structures with SparseMeXt on Nuscenes \texttt{val} set.}
\label{table:FPN}
 \begin{tabular}{c c c |c c c c} 
 \toprule
\multirow{2}*{Type} & \multirow{2}*{Flops} & \multirow{2}*{Params} & \multicolumn{4}{c}{$AP$} \\
& & & $ped.$ & $div.$ & $bou.$ & $avg.$ \\
 \midrule
MiMo & 193.6 & 85.8 & 53.4 & 58.9 & 60.0 & 57.4 \\ 
 SiMo & 96.0 & 39.7 & \textbf{57.5} & \textbf{63.4} & \textbf{64.2} & \textbf{61.7}(\textcolor{OliveGreen}{+4.25}) \\
 
 \bottomrule
 \end{tabular}

\end{table}

\subsubsection{Changing stage compute ratio}
In SparseDrive, a no-temporal perception with one stage and a temporal fusion perception with five stages are used for network construction. In the 3D object detection task, the diversity of objects benefits from more stages in the temporal fusion module. However, due to the fewer classifications of the map task and the better temporal invariance of geometric positions, having more stages can lead to parameter redundancy and overfitting of the network. We systematically explore the combination of temporal and non-temporal design spaces, as presented in Table~\ref{table: Stage ratid}. The optimal configuration consists of one non-temporal stage decoder and four temporal stage decoders, achieving an additional 0.07\% improvement in average mAP.

\begin{table}[h!]
\centering
\caption{The comparison different stage ratios on Nuscenes \texttt{val} set.}
\label{table: Stage ratid}
 \begin{tabular}{c c |c c c c c } 
\toprule
\multirow{2}*{no-temporal} & 
\multirow{2}*{temporal} & 
\multicolumn{4}{c}{$AP$} \\ 
& & $ped.$ & $div.$ & $bou.$ & $avg.$ \\
 \midrule
 1 & 5 & {54.2} & 58.8 & 59.1 & 57.4 \\
 1 & 4 & 53.4 & 58.9 & \textbf{60.0} & \textbf{57.4} (\textcolor{OliveGreen}{+0.07})\\ 
 1 & 3 & 48.2 & 58.0 & 59.4 & 55.2 (\textcolor{Red}{-2.2})\\
 2 & 4 & 51.2 &  \textbf{59.7} & 58.9 & 56.6 (\textcolor{Red}{-0.8})\\
 
 \bottomrule
 \end{tabular}
\end{table}

 

\subsubsection{Decouple Decoder}
 The inherent conflict between classification and regression tasks has long been a persistent challenge, primarily due to their fundamental differences in feature sensitivity. Although existing methods, such as SparseDrive, have alleviated this issue to some extent by introducing decoupled refinement layers at the instance feature level, performance remains limited due to the inevitable feature conflicts arising during image feature point extraction. 
 
 \begin{figure}[h!] 
    \centering
    \includegraphics[width=0.5\textwidth]{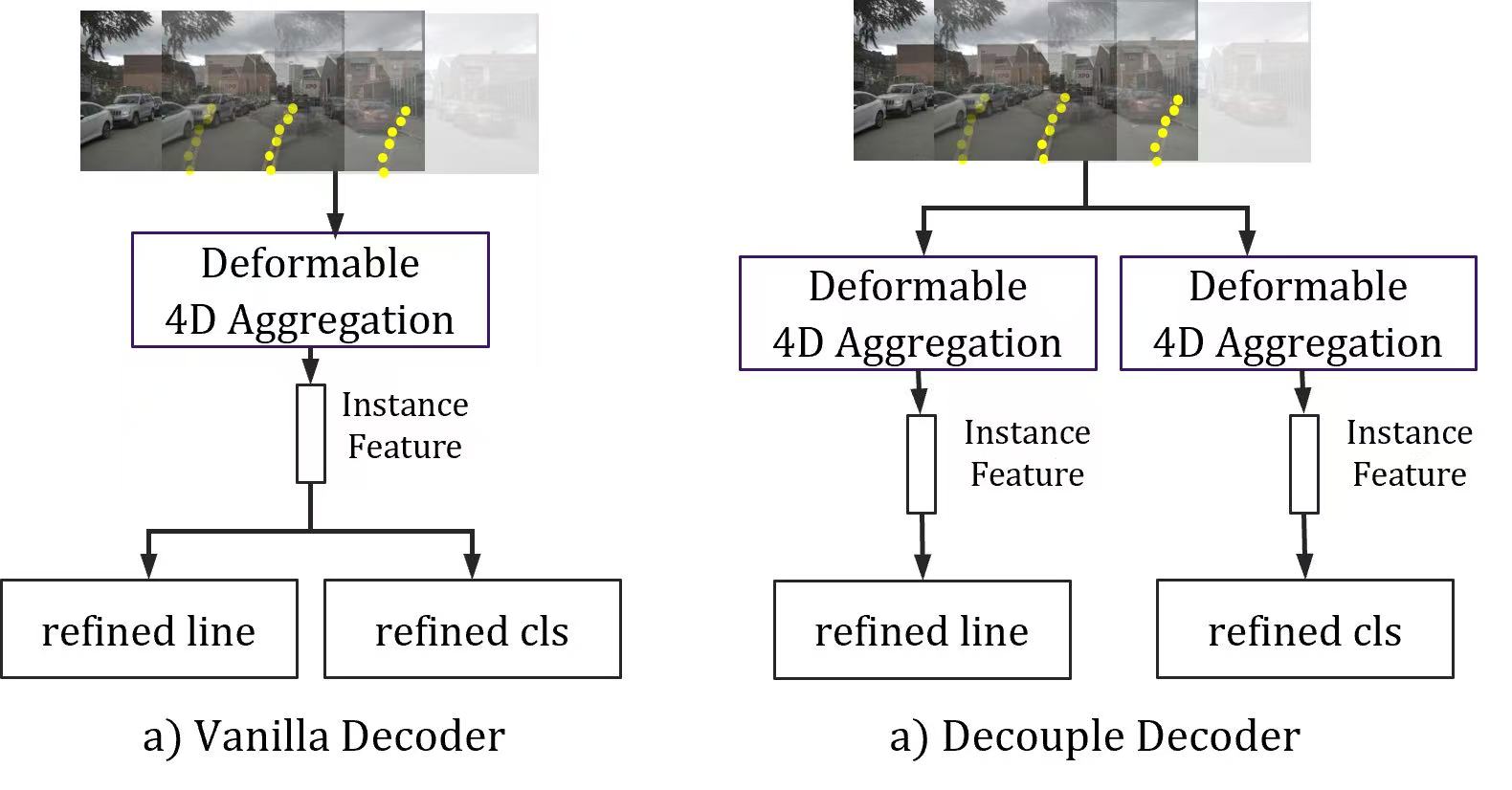} 
    \caption{Architecture of the decouple decoder. We independently split decoder to two branch.}
    \label{fig:decoupledecoder} 
\end{figure}


To address the inherent conflict between classification and regression tasks in object detection, we propose a novel task-decoupling method based on deformable feature aggregation layers. As illustrated in Fig.\ref{fig:decoupledecoder}, this innovative approach constructs independent feature sampling points for classification and regression tasks during the feature extraction phase, effectively achieving feature space decoupling between these two fundamental tasks. The method employs a dual-branch architecture where deformable convolution layers are strategically implemented to adaptively sample task-specific features from different spatial locations. For classification tasks, the sampling points focus on regions rich in categorical information, while for regression tasks, they concentrate on areas crucial for precise boundary localization. This spatial separation of feature extraction enables each task-specific branch to optimize its performance without mutual interference. Experimental results in Table~\ref{table: Decoupledecoder} demonstrates that the proposed method significantly improves detection performance across multiple benchmarks, achieving a notable 0.6\% increase in mAP compare with vanilla decoupled approaches.

\begin{table}[ht]
\centering
\caption{The comparison of Different decouple types on Nuscenes \texttt{val} set.}
\label{table: Decoupledecoder}
 \begin{tabular}{c| c c c c } 
 \toprule
\multirow{2}*{Decouple Type} & \multicolumn{4}{c}{$AP$} \\
& $ped.$ & $div.$ & $bou.$ & $avg.$ \\
 \midrule
 Baseline & 57.5 & 63.4 & \textbf{64.2} & 61.7 \\ 
 Decouple-DFA  & \textbf{59.1} & \textbf{64.3} & 63.4 & \textbf{62.3} \textcolor{OliveGreen}{(+0.6)} \\

 \bottomrule
 \end{tabular}
\end{table}

\subsection{Instance and scene auxiliary segmentation}
\begin{figure}[htb] 
    \centering
    \includegraphics[width=0.5\textwidth]{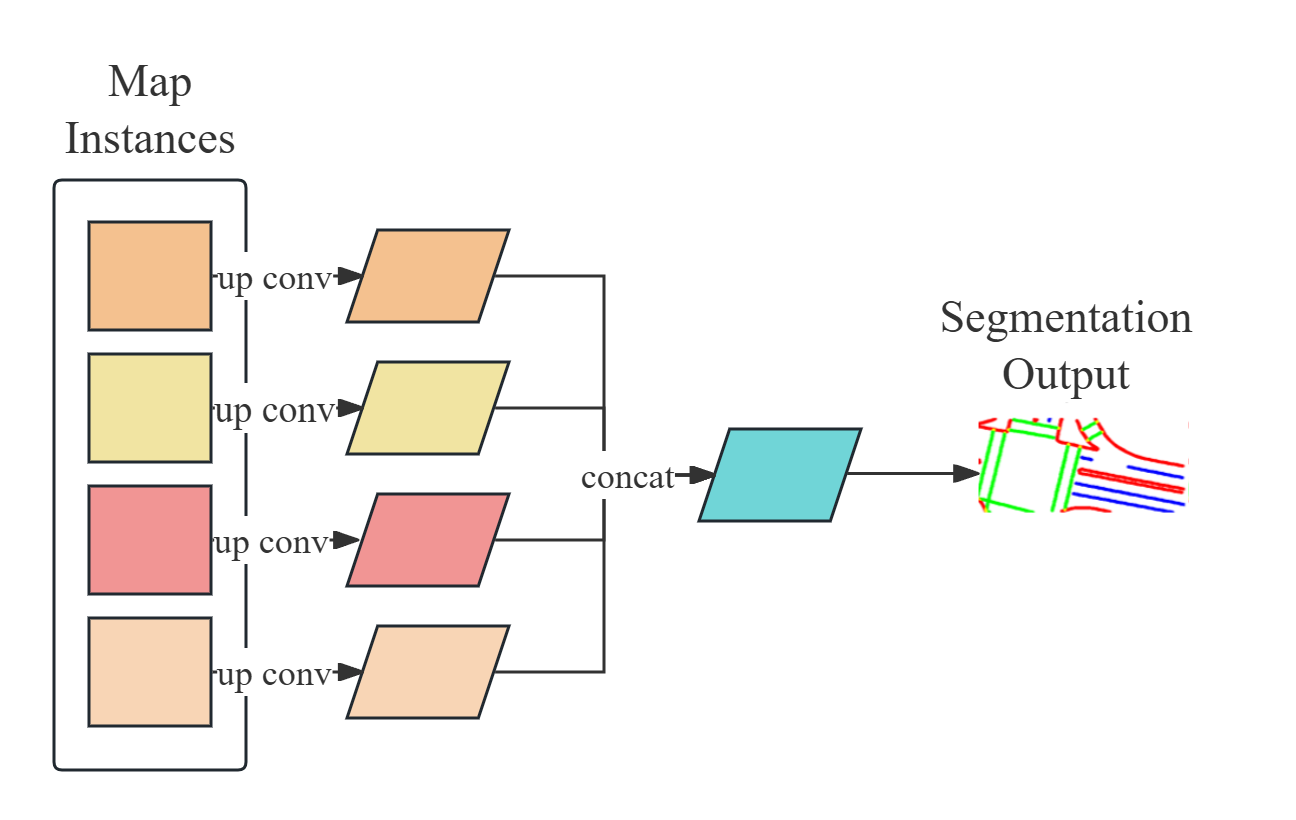} 
    \caption{Architecture of the sparse-dense segmentation head. }
    \label{fig:sparse-dense-seg} 
\end{figure}

MapTRv2\cite{liao2024maptrv2endtoendframeworkonline} introduces auxiliary foreground segmentation on bird’s eye view (BEV), and leverage depth supervision to guide the backbone to learn 3D geometrical information. However, the auxiliary foreground segmentation task relies on an explicit dense BEV space, which is somewhat contradictory to the design philosophy of sparse architectures. We first build a parallel dense BEV space based on a sparse architecture for auxiliary foreground segmentation supervision tasks. However, we find that this not only increased the number of training parameters significantly, but also did not yield any noticeable performance improvement. To address this, as shown in Fig.\ref{fig:sparse-dense-seg}, we design a query-centric sparse-dense reconstruction module to build the complementary auxiliary foreground segmentation task, enabling the network to obtain global foreground supervision.

\begin{table}[htb]
\centering
\caption{The comparison between with and without Segmentation Loss on Nuscenes \texttt{val} set.}
\label{table: Seg Loss}
 \begin{tabular}{c| c c c c } 
 \toprule
\multirow{2}*{Auxiliary Loss Type} & \multicolumn{4}{c}{$AP$} \\
& $ped.$ & $div.$ & $bou.$ & $avg.$ \\
 \midrule
 Baseline(w/o seg loss) & 59.3 & \textbf{66.7} & 65.4 & 63.8 \\ 
 with seg loss  & \textbf{61.6} & 66.5 & \textbf{66.1} & \textbf{64.7} \textcolor{OliveGreen}{(+0.9)} \\
 
 \bottomrule
 \end{tabular}
\end{table}

For detail, this auxiliary segmentation head takes instance features from a SparseMeXt as input and transforms them into a BEV dense representation using up-convolution layers. Then multiple instance features are concatenated to integrate spatial and contextual information. The network ultimately outputs a segmentation map, providing a pixel-wise classification of the input map. Notably, this part of the network is disabled during inference, ensuring no additional computational overhead is introduced in the deployment phase. As Table~\ref{table: Seg Loss} shows, the model with our proposal query-centric sparse-dense auxiliary segmentation task achieving a further 0.9\% improvement in mAP.

\subsection{Physical prior query denoising (PPDN)}
DN-DETR feeds ground-truth bounding boxes with noise into the transformer decoder and trains the model to reconstruct the original boxes. However, unlike directly adding noise to the position ($x,y,z$), width($\mathrm{d}x$), and height($\mathrm{d}y$) of a bounding box, for curve data formats, the transformer decoder learns mapping relationship between ground truth with noise and ground truth by randomly adding noise to each point is detrimental to model convergence. By leveraging physical prior knowledge about the static component and geometry, we design four types of noise patterns for map elements, shown in Fig.\ref{fig:Overview}, rotation noise, location noise, scale noise and curvature noise. \textbf{rotation noise}: Assuming the average point of all points on a line as the anchor points, and using this anchor point as the origin of a Cartesian coordinate system, random rotational noise $\theta$ is added. \textbf{location noise}: Based on the aforementioned anchor point, add same noise in the $x$ and $y$ directions to all points on the line. \textbf{scale noise}: Apply random scaling noise to the $x$ and $y$ coordinates of all points on the line. \textbf{curvature noise}: We determine lane curvature ($div.$ and $bou.$) by computing the second order derivatives as the difference of tangents at consecutive points divided by their euclidean distance. After adding noise based on the curvature, the displacement transformed by the curvature is adjusted to each point. Unlike Denoising on object detection task, the positions and categorical relationships of road static elements exhibit strong relative dependencies, so we do not adopt categorical noise. Table~\ref{table: PPDN} demonstrates our PPDN can effectively enhance performance by achieving a 0.5\% improvement in mAP.

\begin{table}[h!]
\centering
\caption{The comparison of models with and without Physical prior query denoising on Nuscenes \texttt{val} set.}
\label{table: PPDN}
 \begin{tabular}{c| c c c c } 
 \toprule
\multirow{2}*{Denoising Type} & \multicolumn{4}{c}{$AP$} \\
& $ped.$ & $div.$ & $bou.$ & $avg.$ \\
 \midrule
 Baseline(w/o PPDN) & 61.6 & 66.5 & 66.1 & 64.7 \\ 
 with PPDN  & \textbf{62.6} & \textbf{67.0} & \textbf{66.1} & \textbf{65.2} \textcolor{OliveGreen}{(+0.5)} \\

 \bottomrule
 \end{tabular}
\end{table}

\section{Experiments}

\subsection{Experimental Settings}

\subsubsection{Dataset}
SparseMeXt is evaluated on the NuScenes map dataset, which consists of two line-shaped map classes (lane divider and road boundary) and one polygon-shaped map class (pdedstrain crossing). We follow the official 700/150/150 split for training/validation/testing scenes to training and testing our SparseMeXt model. Basic NuScenes map dataset detection range is 30m and 60m along the x and y axes. In our experiments, we additionally generate a long-range NuScenes map dataset, extending the ranges of x and y to 60m and 90m, respectively. 


\subsubsection{Training Implementation Details} In general, we primarily follow SparseDrive’s training protocol. The resolution of source images is The input image resolution is set to 704 pixels wide by 256 pixels high. All model architectures are implemented
with the PyTorch library and the training is distributed on 4 NVIDIA A100 GPU devices with
Automated Mixed Precision. The mini-batch size per device is set to 16. The training period lasts for 100 epochs for system-level comparison. The initial learning rate is 0.006 and is
decayed following a half-cosine-shaped function. The learning rate of the backbone is multiplied by
a factor of 1/10 because it has been pre-trained. We find that the final result is robust to the initial learning rate, thanks to the AdamW optimizer. The weight decay is fixed as 0.01 and
the $l2$ norm of gradients is clipped to be no more than 35. The perception range is [-15.0m, 15.0m] along the X-axis and [-30.0m, 30.0m] along the Y-axis with reference to the ego-vehicle for base range experiments and [-30.0m, 30.0m ] along the X-axis and [-45.0m, 45.0m] along the Y-axis with reference to the ego-vehicle for long-range experiments.
\subsubsection{Inference Implementation Details} The inference process is straightforward. For given surrounding images, the model directly predicts 100 map elements along with their corresponding confidence scores. These scores reflect the reliability of each predicted map element, allowing the top-scoring predictions to be used immediately without additional post-processing. The inference time is evaluated on a single NVIDIA GeForce RTX 3090 GPU with a batch size of 1.

\subsection{Quantitative Results}
We evaluate the performance of our proposed SparseMeXt series on the nuScenes dataset and compare it against the state-of-the-art methods, including HDMapNet, VectorMapNet, MapTR, MapTRv2, MapNeXt, and SparseDrive. The results are summarized in Table~\ref{tab:main}, which reports the Average Precision (AP) for pedestrian crossings ($ped.$), lane dividers ($div.$), and road boundaries ($bou.$), as well as the average AP and inference speed (FPS) on both RTX3090 and A100.

Our SparseMeXt-Tiny model, with a ResNet-18 backbone, achieves an mAP of 55.5\%, outperforming MapTR-Nano (45.9\%) and MapTRv2-Tiny (52.3\%) by a significant margin. With a larger backbone, SparseMeXt-Base (ResNet-50) further improves the mAP to 65.2\%, demonstrating superior performance over R50-based MapTR-Tiny (58.7\%) and MapTRv2-Base (63.0\%). In particular, SparseMeXt-Large (ResNet-101) achieves the highest mAP of 68.9\%, surpassing all competing methods, including MapTRv2-Base (68.7\%) and SparseDrive-B (56.2\%).

\begin{table*}[ht]
	\caption{State-of-the-art comparison on nuScenes \texttt{val} set.}
	\label{tab:main}
	\centering
	\resizebox{\linewidth}{!}{
		\begin{tabular}{l|ccc|cccc|cc}
			\toprule
			\multirow{2}*{Architecture}  & \multirow{2}*{Modality} & \multirow{2}*{Backbone}  & \multirow{2}*{Epochs} & \multicolumn{4}{c|}{$AP$} &  \multicolumn{2}{c}{FPS} \\
			& & & & $ped.$ & $div.$ & $bou.$ & $avg.$ & RTX3090 & A100 \\
			\midrule
			HDMapNet~\cite{li2022hdmapnetonlinehdmap} & C & EffNet-B0  & 30 & 14.4 & 21.7 & 33.0 & 23.0 & 0.8 & -   \\
			\midrule
			VectorMapNet~\cite{liu2023vectormapnetendtoendvectorizedhd} & C & R50 & 110 & 36.1 & 47.3 & 39.3 & 40.9 & 2.9 & - \\
			\midrule
			MapTR-Nano~\cite{liao2023maptrstructuredmodelinglearning} & C & R18 & 110 & 39.6 & 49.9 & 48.2 & 45.9 & 29.2 & 50.5 \\
			MapTR-Tiny~\cite{liao2023maptrstructuredmodelinglearning} & C & R50 & 110 & 56.2 & 59.8 & 60.1 & 58.7 & 12.6 & 20.0 \\

            \midrule
            MapTRv2-Tiny~\cite{liao2024maptrv2endtoendframeworkonline} & C & R18 & 110 & 46.9 & 55.1 & 54.9 & 52.3 & 33.7 & - \\
            MapTRv2-Base~\cite{liao2024maptrv2endtoendframeworkonline} & C & R50 & 110 & 68.1 & 68.3 & 69.7 & 68.7  & 14.1 & - \\
                        
            \midrule
            MapNeXt-Tiny \cite{li2024mapnext} & C & R50 & 110 & 57.7 & 65.3 & 65.8 & 63.0  & 12.7 & 20.3 \\
            \midrule 
            SparseDrive-S \cite{sun2024sparsedriveendtoendautonomousdriving} & C & R50 & 100 & 49.9 & 57.0 & 58.4 & 55.1  & 18.0 & - \\
            SparseDrive-B \cite{sun2024sparsedriveendtoendautonomousdriving} & C & R101 & 100 & 53.2 & 56.3 & 59.1 & 56.2  & 14.3 & - \\    
            \midrule 
            SparseMeXt-Tiny (Our) & C & R18 & 100 & 51.4 & 58.2 & 57.0 & 55.5  & 32.9 & - \\
            SparseMeXt-Base (Our) & C & R50 & 100 & 62.6 & 67.0 & 66.1 & 65.2  & 25.4 & - \\
            SparseMeXt-Large (Our) & C & R101 & 100 & \textbf{66.4} & \textbf{70.8} & \textbf{69.9} & \textbf{68.9}  & 20.2 & - \\
			\bottomrule
		\end{tabular}
}
\end{table*}

In terms of computational efficiency, SparseMeXt-Tiny achieves an impressive 32.9 FPS on RTX3090, making it highly suitable for real-time applications. The larger variants, SparseMeXt-Base and SparseMeXt-Large, maintain competitive speeds of 25.4 FPS and 20.2 FPS, respectively, while delivering state-of-the-art accuracy.

In addition, we include the centerline learning task in the SparseMaXt, as it provides direction and connectivity for downstream motion prediction and planning, which SparseDrive-map does not account for. As shown in Table~\ref{table: center-line}, SparseMeXt achieves 58.5\% mAP on the Nuscenes dataset. Extending SparseMeXt to centerline paves the way for end-to-end planning.

\begin{table}[ht]
\centering
\caption{The performance comparison between models on centerline task on Nuscenes \texttt{val} set.}
\label{table: center-line}
\begin{tabular}{c@{\hskip 5pt}|c@{\hskip 5pt}|ccccc}
\toprule
\multirow{2}*{Architecture} & \multirow{2}*{Epoch} & \multicolumn{5}{c}{$AP$} \\
 & & $ped.$ & $div.$ & $bou.$ & $cen.$ & $avg.$ \\
 \midrule
MapTRv2(R50)          & 110 & 50.1          & 53.9          & 58.8            & 53.1            & 54.0  \\ \midrule
SparseMeXt-Tiny(R18)  & 100 & 47.3          & 60.7          & 57.9            & 52.0            & 54.5   \\
SparseMeXt-Base(R50)  & 100 & \textbf{51.7} & \textbf{64.6} & \textbf{62.3}   & \textbf{56.8}   & \textbf{58.8}   \\

\bottomrule
\end{tabular}
\end{table}

To better meet the long-range perception requirements of autonomous driving, we have extend the perception range to -30m to 30m meters laterally and -45m to 45m meters longitudinally. As shown in Table~\ref{table: long-range}, we evaluate the performance of our proposed SparseMeXt series at a 90m perception range on the nuScenes dataset and compare it with MapTR. At a 60x90m range, SparseMeXt-Base achieves 47.6\% mAP, it shows a 7.4\% mAP significant improvement over MapTR.

\begin{table}[ht]
\centering
\caption{The performance comparison of long range HD map construction on Nuscenes \texttt{val} set.}
\label{table: long-range}
\begin{tabular}{c@{\hskip 5pt}|c@{\hskip 5pt}|cccc}
\toprule
\multirow{2}*{Architecture} & \multirow{2}*{Range} & \multicolumn{4}{c}{$AP$} \\
 & & $ped.$ & $div.$ & $bou.$ & $avg.$ \\
\midrule

$90\times60m$ & MapTR(R50)      & 46.3          & 36.3           & 38.0           & 40.2 \\ 
$90\times60m$ & SparseMeXt-Tiny(R18) & 36.4          & 43.3           & 35.7           & 38.5 \\
$90\times60m$ & SparseMeXt-Base(R50) & \textbf{46.3} & \textbf{49.6}  & \textbf{46.6}  & \textbf{47.6} \\
\bottomrule
\end{tabular}
\end{table}

\subsection{Ablation Experiments}
The SparseDrive Baseline (only training in map task) achieves an mAP of 57.4\%, with AP scores of 54.2\%, 58.8\%, and 59.1\% for pedestrian crossings ($ped.$), lane dividers ($div.$), and road boundaries ($bou.$), respectively. This serves as the starting point for our ablation study. 
In Tabel\ref{table: Ablation Experiments} we show how we build SparseMeXt on SparseDrive. We first replace the backbone, decoder stage radio and single-in-multiple-out FPN. Then we gradually add the auxiliary sparse-dense supervision, decouple deformable feature aggregation, and PPDN. With all these components, SparseMeXt-Base (ResNet50) achieves 65.2\% mAP, which is 10.1\% mAP higher than SparseDrive with 14 fps faster.

\begin{table}[ht]
\centering
\caption{Ablation Experiments.}
\label{table: Ablation Experiments}
 \begin{tabular}{l| c c c |l } 
 \toprule
 \multirow{2}*{Cost Type} & \multicolumn{3}{c|}{$AP$} & \multirow{2}*{mAP} \\
& $ped.$ & $div.$ & $bou.$ \\
 \midrule
SparseDrive-Map Baseline & 54.2 & 58.8 & 59.1 & 57.4 \\ 
\hspace{1em}+ Different stage radio & 53.4 & 59.0 & 60.0 & 57.4 \textcolor{OliveGreen}{(+0.07)} \\
\hspace{1em}+ SIMO & 57.5 & 63.4 & 64.2 & 61.7 \textcolor{OliveGreen}{(+4.25)} \\
\hspace{1em}+ Decouple-DFA & 59.1 & 64.3 & 63.4 & 62.3 \textcolor{OliveGreen}{(+0.6)} \\ 
\hspace{1em}+ Pretraining Backbone & 59.3 & 66.7 & 65.4 & 63.8 \textcolor{OliveGreen}{(+1.5)} \\
\hspace{1em}+ Auxiliary segmentation loss & 61.6 & 66.5 & 66.1 & 64.7 \textcolor{OliveGreen}{(+0.9)} \\
\hspace{1em}+ PPDN & 62.6 & 67.0 & 66.1 & 65.2 \textcolor{OliveGreen}{(+0.5)}\\
 \bottomrule
 \end{tabular}
\end{table}

\subsection{Qualitative Results}
\begin{figure*}[htb] 
    \centering
    \includegraphics[width=1.0\textwidth]{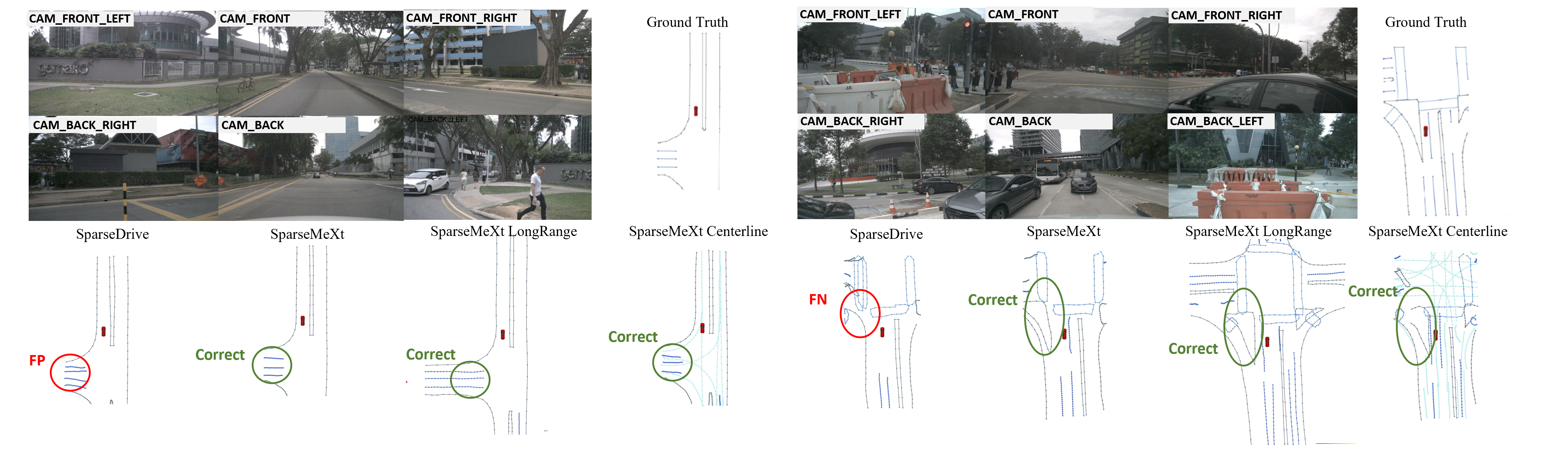} 

    \caption{Qualitative results of SparseDrive, SparseMeXt, SparseMext LongRange and SparseMeXt Centerline.}
    \label{fig:vis_1} 
\end{figure*}

We present the predicted vectorized HD map results for the nuScenes dataset in Fig.~\ref{fig:vis_1}. The SparseMeXt series of models not only demonstrates impressive performance in complex driving scenes but also exhibits enhanced capabilities in long-range scenarios and centerline detection.

\section{Conclusion}
In conclusion, SparseMeXt represents a significant step forward in the field of online HD map construction on the sparse paradigm, offering a robust and efficient solution for autonomous driving systems. 

Our contributions are threefold. First, we introduce a network architecture optimized for sparse map feature extraction, enabling more effective feature aggregation and representation learning for online HD map construction. Second, our sparse-to-dense auxiliary segmentation supervision method compensates for the limitations of sparse BEV features by enhancing semantic and geometric understanding, thereby significantly improving HD map reconstruction performance. Third, the proposed physical prior-based query denoising (PPDN) training strategy, which generates various types of physical prior noise, significantly improves prediction stability and robustness in online HD map construction. We believe that our work will inspire further research and innovation in this critical area, ultimately contributing to the development of safer and more reliable autonomous vehicles.







\section{ACKNOWLEDGMENT}
This work is supported in the THU-Bosch JCML. We would like to acknowledge our friends Jingyu Zhang, Prof.Jian Pu and Prof.Liangyao Chen from Fudan University, Prof.Osamu Yoshie from Waseda University for their fruitful discussions and follow-ups.



\bibliographystyle{IEEEtran}
\bibliography{reference}

\begin{thebibliography}{10}
\providecommand{\url}[1]{#1}
\csname url@samestyle\endcsname
\providecommand{\newblock}{\relax}
\providecommand{\bibinfo}[2]{#2}
\providecommand{\BIBentrySTDinterwordspacing}{\spaceskip=0pt\relax}
\providecommand{\BIBentryALTinterwordstretchfactor}{4}
\providecommand{\BIBentryALTinterwordspacing}{\spaceskip=\fontdimen2\font plus
\BIBentryALTinterwordstretchfactor\fontdimen3\font minus \fontdimen4\font\relax}
\providecommand{\BIBforeignlanguage}[2]{{%
\expandafter\ifx\csname l@#1\endcsname\relax
\typeout{** WARNING: IEEEtran.bst: No hyphenation pattern has been}%
\typeout{** loaded for the language `#1'. Using the pattern for}%
\typeout{** the default language instead.}%
\else
\language=\csname l@#1\endcsname
\fi
#2}}
\providecommand{\BIBdecl}{\relax}
\BIBdecl

\bibitem{zhang2014loam}
J.~Zhang, S.~Singh \emph{et~al.}, ``Loam: Lidar odometry and mapping in real-time.'' in \emph{Robotics: Science and systems}, vol.~2, no.~9.\hskip 1em plus 0.5em minus 0.4em\relax Berkeley, CA, 2014, pp. 1--9.

\bibitem{shan2018legoloam}
T.~Shan and B.~Englot, ``Lego-loam: Lightweight and ground-optimized lidar odometry and mapping on variable terrain,'' in \emph{2018 IEEE/RSJ International Conference on Intelligent Robots and Systems (IROS)}.\hskip 1em plus 0.5em minus 0.4em\relax IEEE, 2018, pp. 4758--4765.

\bibitem{bai2022fasterlio}
C.~Bai, T.~Xiao, Y.~Chen, H.~Wang, F.~Zhang, and X.~Gao, ``Faster-lio: Lightweight tightly coupled lidar-inertial odometry using parallel sparse incremental voxels,'' \emph{IEEE Robotics and Automation Letters}, vol.~7, no.~2, pp. 4861--4868, 2022.

\bibitem{xue2022legoloamsc}
G.~Xue, J.~Wei, R.~Li, and J.~Cheng, ``Lego-loam-sc: An improved simultaneous localization and mapping method fusing lego-loam and scan context for underground coalmine,'' \emph{Sensors}, vol.~22, no.~2, p. 520, 2022.

\bibitem{zheng2022fastlivo}
C.~Zheng, Q.~Zhu, W.~Xu, X.~Liu, Q.~Guo, and F.~Zhang, ``Fast-livo: Fast and tightly-coupled sparse-direct lidar-inertial-visual odometry,'' in \emph{2022 IEEE/RSJ international conference on intelligent robots and systems (IROS)}.\hskip 1em plus 0.5em minus 0.4em\relax IEEE, 2022, pp. 4003--4009.

\bibitem{reinauer2021persformer}
R.~Reinauer, M.~Caorsi, and N.~Berkouk, ``Persformer: A transformer architecture for topological machine learning,'' \emph{arXiv preprint arXiv:2112.15210}, 2021.

\bibitem{zhou2022crossviewlane}
B.~Zhou and P.~Kr{\"a}henb{\"u}hl, ``Cross-view transformers for real-time map-view semantic segmentation,'' in \emph{Proceedings of the IEEE/CVF conference on computer vision and pattern recognition}, 2022, pp. 13\,760--13\,769.

\bibitem{can2021structured}
Y.~B. Can, A.~Liniger, D.~P. Paudel, and L.~Van~Gool, ``Structured bird's-eye-view traffic scene understanding from onboard images,'' in \emph{Proceedings of the IEEE/CVF International Conference on Computer Vision}, 2021, pp. 15\,661--15\,670.

\bibitem{huang2021bevdet}
J.~Huang, G.~Huang, Z.~Zhu, Y.~Ye, and D.~Du, ``Bevdet: High-performance multi-camera 3d object detection in bird-eye-view,'' \emph{arXiv preprint arXiv:2112.11790}, 2021.

\bibitem{li2022bevformer}
Z.~Li, W.~Wang, H.~Li, E.~Xie, C.~Sima, T.~Lu, Y.~Qiao, and J.~Dai, ``Bevformer: Learning bird’s-eye-view representation from multi-camera images via spatiotemporal transformers,'' in \emph{European conference on computer vision}.\hskip 1em plus 0.5em minus 0.4em\relax Springer, 2022, pp. 1--18.

\bibitem{li2023bevdepth}
Y.~Li, Z.~Ge, G.~Yu, J.~Yang, Z.~Wang, Y.~Shi, J.~Sun, and Z.~Li, ``Bevdepth: Acquisition of reliable depth for multi-view 3d object detection,'' in \emph{Proceedings of the AAAI Conference on Artificial Intelligence}, vol.~37, no.~2, 2023, pp. 1477--1485.

\bibitem{li2022hdmapnetonlinehdmap}
\BIBentryALTinterwordspacing
Q.~Li, Y.~Wang, Y.~Wang, and H.~Zhao, ``Hdmapnet: An online hd map construction and evaluation framework,'' 2022. [Online]. Available: \url{https://arxiv.org/abs/2107.06307}
\BIBentrySTDinterwordspacing

\bibitem{liu2023vectormapnetendtoendvectorizedhd}
\BIBentryALTinterwordspacing
Y.~Liu, T.~Yuan, Y.~Wang, Y.~Wang, and H.~Zhao, ``Vectormapnet: End-to-end vectorized hd map learning,'' 2023. [Online]. Available: \url{https://arxiv.org/abs/2206.08920}
\BIBentrySTDinterwordspacing

\bibitem{liao2023maptrstructuredmodelinglearning}
\BIBentryALTinterwordspacing
B.~Liao, S.~Chen, X.~Wang, T.~Cheng, Q.~Zhang, W.~Liu, and C.~Huang, ``Maptr: Structured modeling and learning for online vectorized hd map construction,'' 2023. [Online]. Available: \url{https://arxiv.org/abs/2208.14437}
\BIBentrySTDinterwordspacing

\bibitem{liao2024maptrv2endtoendframeworkonline}
\BIBentryALTinterwordspacing
B.~Liao, S.~Chen, Y.~Zhang, B.~Jiang, Q.~Zhang, W.~Liu, C.~Huang, and X.~Wang, ``Maptrv2: An end-to-end framework for online vectorized hd map construction,'' 2024. [Online]. Available: \url{https://arxiv.org/abs/2308.05736}
\BIBentrySTDinterwordspacing

\bibitem{carion2020endtoendobjectdetectiontransformers}
\BIBentryALTinterwordspacing
N.~Carion, F.~Massa, G.~Synnaeve, N.~Usunier, A.~Kirillov, and S.~Zagoruyko, ``End-to-end object detection with transformers,'' 2020. [Online]. Available: \url{https://arxiv.org/abs/2005.12872}
\BIBentrySTDinterwordspacing

\bibitem{sun2024sparsedriveendtoendautonomousdriving}
\BIBentryALTinterwordspacing
W.~Sun, X.~Lin, Y.~Shi, C.~Zhang, H.~Wu, and S.~Zheng, ``Sparsedrive: End-to-end autonomous driving via sparse scene representation,'' 2024. [Online]. Available: \url{https://arxiv.org/abs/2405.19620}
\BIBentrySTDinterwordspacing

\bibitem{chen2022persformer3dlanedetection}
\BIBentryALTinterwordspacing
L.~Chen, C.~Sima, Y.~Li, Z.~Zheng, J.~Xu, X.~Geng, H.~Li, C.~He, J.~Shi, Y.~Qiao, and J.~Yan, ``Persformer: 3d lane detection via perspective transformer and the openlane benchmark,'' 2022. [Online]. Available: \url{https://arxiv.org/abs/2203.11089}
\BIBentrySTDinterwordspacing

\bibitem{liu2022petrv2unifiedframework3d}
\BIBentryALTinterwordspacing
Y.~Liu, J.~Yan, F.~Jia, S.~Li, A.~Gao, T.~Wang, X.~Zhang, and J.~Sun, ``Petrv2: A unified framework for 3d perception from multi-camera images,'' 2022. [Online]. Available: \url{https://arxiv.org/abs/2206.01256}
\BIBentrySTDinterwordspacing

\bibitem{li2022dn}
F.~Li, H.~Zhang, S.~Liu, J.~Guo, L.~M. Ni, and L.~Zhang, ``Dn-detr: Accelerate detr training by introducing query denoising,'' in \emph{Proceedings of the IEEE/CVF conference on computer vision and pattern recognition}, 2022, pp. 13\,619--13\,627.

\bibitem{yuan2024streammapnet}
T.~Yuan, Y.~Liu, Y.~Wang, Y.~Wang, and H.~Zhao, ``Streammapnet: Streaming mapping network for vectorized online hd map construction,'' in \emph{Proceedings of the IEEE/CVF Winter Conference on Applications of Computer Vision}, 2024, pp. 7356--7365.

\bibitem{xu2021confine}
G.~Xu, S.~Tang, Z.~Yu, and K.~Fu, ``Confine keypoint triplets for object detection,'' in \emph{2021 IEEE International Conference on Artificial Intelligence and Industrial Design (AIID)}.\hskip 1em plus 0.5em minus 0.4em\relax IEEE, 2021, pp. 608--613.

\bibitem{liu2016ssd}
W.~Liu, D.~Anguelov, D.~Erhan, C.~Szegedy, S.~Reed, C.-Y. Fu, and A.~C. Berg, ``Ssd: Single shot multibox detector,'' in \emph{Computer Vision--ECCV 2016: 14th European Conference, Amsterdam, The Netherlands, October 11--14, 2016, Proceedings, Part I 14}.\hskip 1em plus 0.5em minus 0.4em\relax Springer, 2016, pp. 21--37.

\bibitem{ren2016faster}
S.~Ren, K.~He, R.~Girshick, and J.~Sun, ``Faster r-cnn: Towards real-time object detection with region proposal networks,'' \emph{IEEE transactions on pattern analysis and machine intelligence}, vol.~39, no.~6, pp. 1137--1149, 2016.

\bibitem{tan2020efficientdet}
M.~Tan, R.~Pang, and Q.~V. Le, ``Efficientdet: Scalable and efficient object detection,'' in \emph{Proceedings of the IEEE/CVF conference on computer vision and pattern recognition}, 2020, pp. 10\,781--10\,790.

\bibitem{tian2022fullyconvolutional}
Z.~Tian, X.~Chu, X.~Wang, X.~Wei, and C.~Shen, ``Fully convolutional one-stage 3d object detection on lidar range images,'' \emph{Advances in Neural Information Processing Systems}, vol.~35, pp. 34\,899--34\,911, 2022.

\bibitem{sunzhigang10592819}
Z.~Sun, Z.~Wang, L.~Halilaj, and J.~Luettin, ``Semanticformer: Holistic and semantic traffic scene representation for trajectory prediction using knowledge graphs,'' \emph{IEEE Robotics and Automation Letters}, vol.~9, no.~9, pp. 7381--7388, 2024.

\bibitem{wang2021detr3d3dobjectdetection}
\BIBentryALTinterwordspacing
Y.~Wang, V.~Guizilini, T.~Zhang, Y.~Wang, H.~Zhao, and J.~Solomon, ``Detr3d: 3d object detection from multi-view images via 3d-to-2d queries,'' 2021. [Online]. Available: \url{https://arxiv.org/abs/2110.06922}
\BIBentrySTDinterwordspacing

\bibitem{lin2022sparse4dv1}
X.~Lin, T.~Lin, Z.~Pei, L.~Huang, and Z.~Su, ``Sparse4d: Multi-view 3d object detection with sparse spatial-temporal fusion,'' \emph{arXiv preprint arXiv:2211.10581}, 2022.

\bibitem{lin2023sparse4dv2}
------, ``Sparse4d v2: Recurrent temporal fusion with sparse model,'' \emph{arXiv preprint arXiv:2305.14018}, 2023.

\bibitem{lin2023sparse4dv3}
X.~Lin, Z.~Pei, T.~Lin, L.~Huang, and Z.~Su, ``Sparse4d v3: Advancing end-to-end 3d detection and tracking,'' \emph{arXiv preprint arXiv:2311.11722}, 2023.

\bibitem{zhang2024sparsead}
D.~Zhang, G.~Wang, R.~Zhu, J.~Zhao, X.~Chen, S.~Zhang, J.~Gong, Q.~Zhou, W.~Zhang, N.~Wang \emph{et~al.}, ``Sparsead: Sparse query-centric paradigm for efficient end-to-end autonomous driving,'' \emph{arXiv preprint arXiv:2404.06892}, 2024.

\bibitem{wang2023objectcentrictemporal}
S.~Wang, Y.~Liu, T.~Wang, Y.~Li, and X.~Zhang, ``Exploring object-centric temporal modeling for efficient multi-view 3d object detection,'' in \emph{Proceedings of the IEEE/CVF International Conference on Computer Vision}, 2023, pp. 3621--3631.

\bibitem{jin2022lookobjects}
\BIBentryALTinterwordspacing
Z.~Jin, D.~Yu, L.~Song, Z.~Yuan, and L.~Yu, ``You should look at all objects,'' 2022. [Online]. Available: \url{https://arxiv.org/abs/2207.07889}
\BIBentrySTDinterwordspacing

\bibitem{chen2021lookonelevelfeature}
\BIBentryALTinterwordspacing
Q.~Chen, Y.~Wang, T.~Yang, X.~Zhang, J.~Cheng, and J.~Sun, ``You only look one-level feature,'' 2021. [Online]. Available: \url{https://arxiv.org/abs/2103.09460}
\BIBentrySTDinterwordspacing

\bibitem{li2024mapnext}
T.~Li, ``Mapnext: Revisiting training and scaling practices for online vectorized hd map construction,'' \emph{arXiv preprint arXiv:2401.07323}, 2024.

\end{thebibliography}

\end{document}